\documentclass[sigconf]{acmart}

\AtBeginDocument{
  }

\setcopyright{acmlicensed}
\copyrightyear{2018}
\acmYear{2018}
\acmDOI{XXXXXXX.XXXXXXX}

\newcommand{\Ls}{\mathcal{L}}

\newcommand{\Modela}{\texttt{CAROL}}
\usepackage{xcolor}

\usepackage{amsfonts}
\usepackage{algorithm}
\usepackage{algpseudocode}
\usepackage{graphicx}
\usepackage{caption}
\usepackage{subcaption}
\usepackage{booktabs}

\usepackage{multirow}
\usepackage{breqn}
\usepackage{arydshln}
\usepackage{xcolor}
\newtheorem{definition}{Definition}

\acmConference[CODS-COMAD Dec'24]{8th Joint International Conference on Data Sciences and Management of Data}{December 18--21,
  2024}{Jodhpur, India}

\begin{document}

\title{Class-Aware Contrastive Optimization for Imbalanced Text Classification}

\author{Grigorii Khvatskii}
\email{gkhvatsk@nd.edu}
\affiliation{
  \institution{University of Notre Dame}
  \city{Notre Dame}
  \state{Indiana}
  \country{USA}
}

\author{Nuno Moniz}
\email{nmoniz2@nd.edu}
\affiliation{
  \institution{University of Notre Dame}
  \city{Notre Dame}
  \state{Indiana}
  \country{USA}
}

\author{Khoa Doan}
\email{doankhoadang@gmail.com}
\affiliation{
  \institution{VinUniversity}
  \city{Hanoi}
  \country{Vietnam}
}

\author{Nitesh V Chawla}
\email{nchawla@nd.edu}
\affiliation{
  \institution{University of Notre Dame}
  \city{Notre Dame}
  \state{Indiana}
  \country{USA}
}

\renewcommand{\shortauthors}{Khvatskii et al.}

\begin{abstract}
The unique characteristics of text data make classification tasks a complex problem. Advances in unsupervised and semi-supervised learning and autoencoder architectures addressed several challenges. However, they still struggle with imbalanced text classification tasks, a common scenario in real-world applications, demonstrating a tendency to produce embeddings with unfavorable properties, such as class overlap. In this paper, we show that leveraging class-aware contrastive optimization combined with denoising autoencoders can successfully tackle imbalanced text classification tasks, achieving better performance than the current state-of-the-art. Concretely, our proposal combines reconstruction loss with contrastive class separation in the embedding space, allowing a better balance between the truthfulness of the generated embeddings and the model's ability to separate different classes. Compared with an extensive set of traditional and state-of-the-art competing methods, our proposal demonstrates a notable increase in performance across a wide variety of text datasets.
\end{abstract}

\ccsdesc[500]{Computing methodologies~Neural networks}
\ccsdesc[500]{Computing methodologies~Learning latent representations}
\ccsdesc[500]{Computing methodologies~Regularization}
\ccsdesc[500]{Computing methodologies~Natural language processing}

\keywords{NLP, Imbalanced Classification, Deep Learning, Transformers, Text Classification}

\maketitle

\section{Introduction}

Natural language processing requires modeling complex dependencies between words in a sentence and sentences of a document, often requiring intricate architectures to achieve both efficacy and efficiency on a variety of tasks. 
Transformer-based model architectures have produced a new state-of-the-art (SOTA) performance on many natural language processing tasks such as machine translation~\cite{raffel2020exploring}, question answering~\cite{reimers2019sentencebert}, text generation~\cite{brown2020language, touvron2023llama}, text-to-image translation~\cite{ramesh2022hierarchical}, as well as a variety of text classification and information extraction tasks~\cite{devlin2019bert, he2021deberta}. Such advances leverage unsupervised and semi-supervised learning to train large language models with little
human intervention. However, current research shows that class imbalance remains a critical and unsolved problem in natural language processing tasks~\cite{henning-etal-2023-survey}.

However, problems such as class imbalance still have a significant impact in model performance in text-based data~\cite{zhu2022sensitivity}, mostly due to embedding configurations with high levels of class overlap.

Imbalanced learning has been a significant hurdle for machine learning tasks for over three decades. Mostly focused on classification tasks in tabular and non-dependency-oriented data, it has received growing attention in other settings: regression~\cite{Ribeiro2020}, time~\cite{Moniz2017, park2023deep}, spatiotemporal forecasting~\cite{oliveira2021biased, gmd-16-251-2023}, and computer vision \cite{dablainDeepSMOTEFusingDeep2021, saini2023tackling}.

The most common strategy to tackle imbalanced learning tasks is data pre-processing, usually leaning on augmentation techniques such as oversampling or data synthetization~\cite{chawlaSMOTESyntheticMinority2002, non-smote-oversampling-comparison}. However, their application in text data is not direct, as no meaningful distance metric will put instances of the same class close together and make classical oversampling possible. Additionally, applying data augmentation techniques to imbalanced datasets may lead to model biases against minority classes~\cite{balestriero2022effects,Carvalho2023} and possibly to direct societal impact~\cite{hovySocialImpactNatural2016}.

\paragraph{Motivation.}
While methods to tackle imbalanced learning on text data have been proposed~\cite{moreoDistributionalRandomOversampling2016,kumarContentBasedBot2021}, they 1) do not take advantage of the latest developments in NLP, 2) often fail in cases where the variance in the minority classes is not enough for sampling with replacement to work or 3) methods based on LLM fine-tuning show that often fail in cases where not enough data is available. 

At the same time, it has been shown that problems such as class imbalance still significantly impact model performance in text-based data~\cite{zhu2022sensitivity}, mostly due to embedding configurations with high levels of class overlap.

Importantly, recent developments in contrastive learning on text \cite{reimers2019sentencebert, henderson2017efficient, gunel2020supervised, suresh2021not} have also not been applied to imbalanced text data, opening an important research opportunity to address the challenge of imbalance classification in text data by combining contrastive learning with LLMs. 

\paragraph{Contribution.}

\begin{figure*}[!h]
    \centering
    \includegraphics[width=\textwidth]{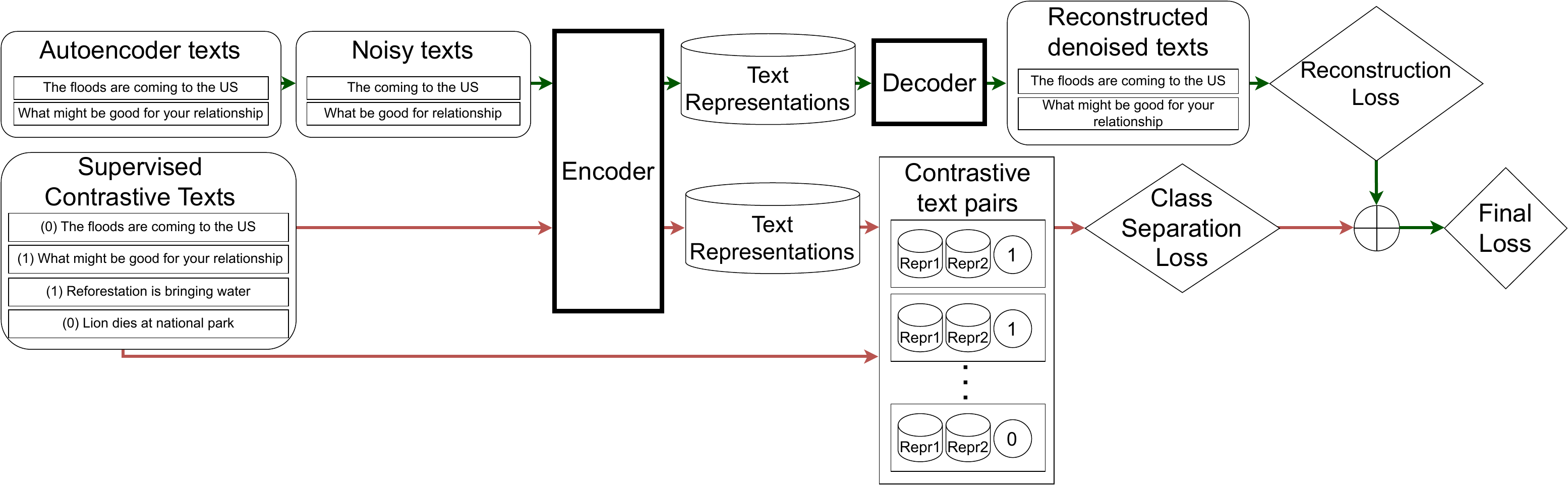}
    \caption{CAROL Training Pipeline}
	\label{CAROLPipeline}
\end{figure*}

In this paper, we present {\Modela}, a \textbf{C}lass-aw\textbf{AR}e
c\textbf{O}ntrastive \textbf{L}oss) that combines a class separation-based loss function with an autoencoder-based model
to tackle imbalanced text binary
classification tasks. {\Modela} is designed to better separate classes in the embedding space
while maintaining the semantic value of the embeddings. 
Our main contributions, beyond the {\Modela} loss function are that {\Modela} allows users to balance reconstruction class separation without losing performance, reaching notable improvements over classical baselines and SOTA approaches. {\Modela} also achieves considerable improvements w.r.t. competing approaches focused on imbalanced text classification tasks \cite{wang2022bert, cao2022adaptable}. We provide further evidence that balancing class separation and reconstruction is a domain-specific problem that should be optimized per domain to achieve optimal downstream classifier performance.

The rest of this paper is organized as follows. Related work is presented in Section~\ref{sec:rw}, focusing on imbalanced and contrastive learning. The architecture of the proposed loss function and the text classification pipeline is presented in Section~\ref{sec:tsa}, followed by an experimental evaluation in Section~\ref{sec:ex}. Finally, in Section~\ref{sec:co}, we discuss our proposal's potential limitations and future improvements.

\section{Related Work}
\label{sec:rw}

In this section, we describe related work on imbalanced domain learning and contrastive learning, focusing on applications to text and other types of unstructured data. We also provide a short overview of unsupervised representation learning from text data.

{\bf{Imbalanced learning. }}
With tabular and structured data, research into imbalanced learning has greatly focused on data augmentation and oversampling. While the simplest approach, random oversampling and its extensions \cite{menardi2014training}, are still widely used, more sophisticated oversampling techniques have received increased attention. Starting with SMOTE \cite{chawlaSMOTESyntheticMinority2002}, these techniques have focused on synthesizing new data examples not previously seen in text data. Over the
years, a large number of extensions of SMOTE were introduced, all addressing different shortcomings of
the original method when used with adverse data conditions \cite{smote-comparison}.
However,
due to the unique characteristics of unstructured data, performing oversampling on such datasets remains a nontrivial task. 
Generative Adversarial Networks were explored for image data as a way for sample generation \cite{douzasEffectiveDataGeneration2018, mullickGenerativeAdversarialMinority2019}. Another approach to image data oversampling is DeepSMOTE \cite{dablainDeepSMOTEFusingDeep2021}, which combines a deep autoencoder with SMOTE for efficient oversampling of images in the embedding space. Specialized imbalanced learning approaches for text data perform feature extraction using bag-of-words, topic modeling, or deep learning-based approaches and then perform oversampling on the extracted features \cite{glazkovaComparisonSyntheticOversampling2020, tahaMULTILABELOVERSAMPLINGUNDERSAMPLING2021, moreoDistributionalRandomOversampling2016}. Cost-based techniques utilizing variations of Focal Loss exist for learning on imbalanced datasets in multiple domains, such as computer vision ~\cite{wang2022bert, cao2022adaptable} and natural language processing \cite{Pasupa_2020}. Surprisingly, research into text encoders robust to class imbalance has received much less attention.

{\bf{Contrastive learning.}}

The formal objective of contrastive learning is to map similar input tensors to output lower-dimensional tensors that are close w.r.t. a given distance metric and to map dissimilar tensors to output vectors that are far away in the output space~\cite{clossieee, clossieee2}.

This can be combined with data augmentation, deriving many positive-negative pairs from relatively small datasets~\cite{chen2020simple}.
Contrastive learning is also used for large language model training. For example, for the original BERT model, one of the pre-training tasks used was next sentence prediction, a type of contrastive learning task \cite{devlin2019bert}. The supervision signal can also be derived from other sources. For example, citation graphs \cite{cohan2020specter} and crowd-sourced question-answer pairs \cite{reimers2019sentencebert, henderson2017efficient} have been utilized in the literature.
Contrastive learning has also been employed for representation learning from unlabeled time series data with good results~\cite{eldele2022self}.

Supervised contrastive learning has also seen some use in the literature. It has been applied to computer vision tasks, where the data used for model training exhibits class overlap. For example, it has been shown that including a label-aware contrastive loss function leads to a marked improvement in performance for various computer vision tasks, \cite{yang2022class}.
Additionally, in computer vision, this loss function was successfully utilized for fault diagnosis, a task well-known for class imbalance and class overlap \cite{zhang2022class}. Contrastive learning was also shown to improve generalization for feature extraction \cite{xie2022domain}.
In computer vision, supervised contrastive learning has also been
employed to improve robustness to class imbalance by disentangling data representations in embedding space \cite{marrakchi2021fighting, choilabel}. Supervised contrastive loss has been used for language model pretraining and has also led to an increase in
performance in noisy conditions and better generalization \cite{gunel2020supervised}, as well as robustness to class overlap \cite{suresh2021not}. Furthermore, a notion of supervised contrastive loss based on distances of samples from different classes was proposed for computer vision tasks \cite{khosla2021supervised}. 

Additionally, supervised contrastive learning was applied for NLP tasks by semi-automatically or manually constructing contrastive pairs of sentences \cite{gao2022simcse}. Utilizing different distances for embedding comparison has proven useful for OOD detection tasks \cite{uppaal-etal-2023-fine}.

{\bf{Representation learning.}}
TSDAE \cite{wangTSDAEUsingTransformerbased2021} is an approach to sentence embedding training using an encoder-decoder construction built from two transformer-based models. 

The loss function used to build the TSDAE model is a denoising autoencoder loss based on cross-entropy, measuring how well original sentences can be reconstructed, defined as follows:

\begin{equation}
	\Ls(\theta) = \mathbb{E}_{x\sim X}[\log P_{\theta}(x|\tilde{x})],
\end{equation}

\noindent where $x$ is an original sentence, $\tilde{x}$ is a sentence with noise added, $X$ is the training text dataset and $\theta$ are the model parameters.

\section{Class-aware contrastive loss}
\label{sec:tsa}

Imbalanced text classification tasks raise several challenges. Particularly, when considering recent advances in natural language processing, we identify two critical ones. Our early experiments show that focusing solely on reconstruction loss may lead to embeddings with a significant class overlap, making classification a challenging task.

However, reconstruction loss does not consider existing class information, focusing only on sentence reconstruction, which may be unnecessary for imbalanced text classification tasks.

Given this, we propose to include information from the instance labels to better inform the relative positions of the instances in the embedding space. An illustration of our proposed training pipeline is shown in Figure \ref{CAROLPipeline}. Our approach is to optimize a metric we call {\it{class separation}}. 

\begin{definition}[Class Separation]
	Class Separation is the combined effect of inter and intraclass distance, where one aims to maximize and minimize the former. This is formally described as follows, leveraging the previous definitions:

	\begin{equation}
		\label{s-eq3}
		S(\theta) = L_{D}(\theta, X_1, X_2) - L_{S}(\theta, X_1, X_2)
	\end{equation}
\end{definition}

This metric consists of 2 sub-metrics: \textit{interclass distance} ($L_{D}\theta$) and \textit{intraclass distance} ($L_{S}\theta$). Please note that averages define interclass and intraclass distances in this context.

\begin{definition}[Interclass Distance]
	Interclass distance is the average distance between pairs of instances of different classes in the embedding space,
	\begin{equation}
		\begin{split}
			\label{s-eq1}
			L_{D}(\theta) = \frac{1}{|X_1||X_2|}
			\sum\limits_{i=1}^{|X_1|}\sum\limits_{j=1}^{|X_2|}D_\theta(\vec x_{1,i}, \vec x_{2,j})
		\end{split}
	\end{equation}

	\noindent where $|X|$ is the number of instances of a class,
	$x_{1,n}$ and $x_{2,n}$ are the instances of classes 1 and 2, respectively, and $D_\theta$ is a distance function between the low-dimensional representations of the points in the embedding space. 
    \newline
\end{definition}

\begin{definition}[Intraclass Distance]
	Intraclass distance is the average distance between pairs of instances of the same class in the embedding space, 
	\begin{equation}
		\begin{split}
			\label{s-eq2}
			L_{S}(\theta) =
			& \frac{1}{|X_1|^2}\sum_{i=1}^{|X_1|}\sum_{j=1}^{|X_1|}D_\theta(\vec x_{1,i}, \vec x_{1,j}) + \\
			& \frac{1}{|X_2|^2}\sum_{i=1}^{|X_2|}\sum_{j=1}^{|X_2|}D_\theta(\vec x_{2,i}, \vec x_{2,j})
		\end{split}
	\end{equation}

	\noindent where $|X|$ is the number of instances of a class,
	$x_{1,n}$ and $x_{2,n}$ are the instances of classes 1 and 2 respectively,
	and $D_\theta$ is a distance function between the representations of points in
	embedding space. 
	
\end{definition}

From Equations \ref{s-eq1} and \ref{s-eq2}, it is evident that the direct computation of this function would incur a $O(n^2)$ computational complexity, which might become prohibitively expensive. However, we can note that since interclass and intraclass distances are defined as averages, we can use sampling to compute their approximations in time bounded by the sample size. To this end, we propose sampling the class instances and then computing the metric using the instances from the sample. From Equation \ref{s-eq3}, it is also evident that $-S(\theta)$ could be used as a loss function, where a smaller value would indicate better class separation. Based on this, we propose the following $\Ls_{\Modela}$ loss function:

\begin{equation}
	\begin{split}
		\label{s-eq4}
		\Ls_{\Modela}(\theta, n) = & L_{S}(\theta, \sigma(X, 1, n), \sigma(X, 2, n)) - \\
		& L_{D}(\theta, \sigma(X, 1, n), \sigma(X, 2, n))
	\end{split}
\end{equation}

\noindent where $D$ is the training dataset, $\theta$ are the model parameters, and $n$ is a hyperparameter of the sample size to use for loss computation. $\sigma(X, c, n)$ is a sampling function that uniformly randomly samples $n$ instances of class $c$ from dataset $X$.

It should be noted that interclass and intraclass distances, and, by extension, class separation, are defined in terms of an opaque distance function. This property makes {\Modela} distance-agnostic: any distance metric can compute the loss function. Such formulation of the loss function also allows it to work without explicit contrastive pair annotations. In our experiments, we analyze the performance of {\Modela} using Euclidean, Chebyshev, and cosine distance metrics.

\subsection{Class-aware Contrastive Loss Computation}

An efficient way to compute $\Ls_{\Modela}$ is presented in Algorithm \ref{CASCLFALGO}. An important concern for computation is that when sampling is used, the number of positive pairs (where classes are the same) will always be smaller than the number of negative pairs (where classes are different). For small sample sizes, this can result in the loss function becoming biased toward negative pairs. To remedy this, we multiply the distances of negative pairs by a small correction coefficient (line 10).

\begin{algorithm*}  
	\caption{$\Ls_{\Modela}(\theta)$}
	\begin{algorithmic}[1]
		\State $\text{Sample Texts}, \text{Sample Labels}\leftarrow$ sample $n$ samples of each class from the training data in random order
		\State $\text{Sample Reprs}\leftarrow$ encode $\text{Sample Texts}$ using the encoder with parameters $\theta$
		\State $loss\leftarrow0$
        \State $m\leftarrow0$
		\For {$Repr1, Repr2, Label1, Label2$ in $Pairs(\text{Sample Reprs}, \text{Sample Labels})$} \Comment{Iterate over all pairs in the sample}
		\State $dist\leftarrow D_\theta(Repr1, Repr2)$ \Comment{Compute distance between representations in the pair}
		\If{$Label1\neq Label2$}
		\State $dist\leftarrow -dist$ \Comment{When the pair is from different classes, no need to apply any correction}
		\Else
		\State $dist\leftarrow dist \cdot (\frac{1}{len(\text{Sample Reprs}) - 1} + 1)$ \Comment{When the pair is from the same class, apply a correction coefficient to avoid bias}
		\EndIf
		\State $loss\leftarrow loss + dist$
        \State $m\leftarrow m + 1$ \Comment{Count the total number of pairs}
		\EndFor
        \State $loss\leftarrow loss / m$ \Comment{Compute the mean class separation}
		\State \Return $loss$
	\end{algorithmic}
	\label{CASCLFALGO}
\end{algorithm*}

The main idea behind the proposed loss function is to include a trade-off between the accuracy of the sentence reconstruction (measured by the cross-entropy loss) and the separation between classes in the data. The proposed loss function also considers the existing supervision signal (class labels) in the data during the training process.

We define the reconstruction loss ($\Ls_{Recon}$) as a cross-entropy loss designed to measure how well the original sentence is reconstructed, leveraging previous work \cite{wangTSDAEUsingTransformerbased2021}:

\begin{equation}
	\Ls_{Recon}(\theta) = \mathbb{E}_{x\sim X}[\log P_{\theta}(x|\tilde{x})]
\end{equation}

The total loss of the autoencoder is then computed as such, where $C$ is a regularization coefficient ranging from 0 to 1:

\begin{equation}
	\Ls_{total}(\theta) = C \cdot \Ls_{\Modela}(\theta) + (1-C) \cdot \Ls_{Recon}(\theta)
\end{equation}

Note that when $C=0$, this loss function becomes equivalent to the simple reconstruction loss.

The $C$ coefficient indicates the trade-off between the faithfulness of the autoencoder reconstruction and class separation in the embedding space. Thus, $\Ls_{total}(\theta)$ loss incorporates both reconstruction loss (responsible for faithfulness) and class separation loss.
This would allow a focus on the faithfulness of the autoencoder in cases where the simple reconstruction loss allows the downstream classifier to perform well while also allowing it to operate in conditions where it might be necessary to
force class separation.

\section{Experiments}
\label{sec:ex}

In this section, we present the experimental results concerning {\Modela}, including details on the datasets and competing methods used. As the main goal for our experimental design, we look to answer the following questions about the performance of our proposed approach:

\begin{enumerate}
	\item Can the embeddings produced by \Modela-based optimization lead to a noticeable improvement in predictive performance?
	\item What is the impact of $C$, and to what extent is it domain-dependent?
	\item What are the weaknesses of our proposed approach, and how are they aligned with limitations from other SOTA methods?
\end{enumerate}

\subsection{Datasets and Methodology}
\label{sec:exds}

\begin{table*}[h]\centering
	\begin{tabular}{lrrrrr}\toprule
		\textbf{Dataset} & \textbf{Train Size} & \textbf{Test Size} & \textbf{Imbalance Ratio} &\textbf{Domain} \\\midrule
		20News \cite{lang95newsweeder} & 11096 & 7370  & 6.15 & Long-form messages\\
		BBC \cite{bbcnewsdataset} & 1903 & 322 & 4.34  & News \\
		CAMEO \cite{Gerner2002ConflictAM} & 1349 & 451 & 4.47  & Events \\
		India Police \cite{halterman-etal-2021-corpus} & 18108 & 3275  & 9.22 & News\\
		InsightCrime \cite{incrime} & 1334 & 319 & 5.12 & News\\
		Spam \cite{spamds} & 4457 & 1115 & 6.46 & Short-form messages\\
		\bottomrule
	\end{tabular}
	\caption{Datasets used for the experimentation}
	\label{DatasetTable}
\end{table*}

We use multiple datasets for validation. The list of the datasets used is provided in Table~\ref{DatasetTable}. We use the versions of the datasets from \cite{hu-etal-2022-conflibert} provided under the GPL-3 license, which is used in a similar setting. All the datasets used the English language.
Given the imbalanced nature of the data, we compute the following metrics: F1 Score, Precision, and Recall measures.
For classical methods and shallow neural networks, the training and evaluation were performed on an Intel Core i7-1255U processor with 40GB of RAM; no GPUs were used. Deep learning and {\Modela}-based methods were trained and evaluated on a server with 4 NVIDIA Quadro RTX 6000 GPUs (96 GB of VRAM), Intel Xeon Gold 6226 CPU, and 188GB of RAM. 

\subsection{Baselines}

We use several baselines to compare the performance of our proposed method, as detailed in the next subsections.

The first group of baselines in our experimental evaluation are more classical methods that do not rely on word embeddings or transformer-based models to work. These include two methods: \textit{i)} the bag-of-words (BoW) and \textit{ii)} TF-IDF. In this work, we use the default scheme in the scikit-learn Python package \cite{scikit-learn}.

All the baselines corresponding to these more classical methods were developed using the SVM~\cite{cortes1995support} learning algorithm. Another baseline approach that we present uses document embeddings. We combined Doc2Vec~\cite{le2014distributed} embeddings, with an SVM-based classifier. The optimal value for the regularization parameter for SVM was selected using 5-fold cross-validation.

A SOTA approach to text classification tasks is BERT with Cross-Entropy loss \cite{devlin2019bert}. For this baseline, we have adapted this model to our tasks by utilizing a pre-trained \texttt{bert-base-uncased} model with mean pooling to produce embeddings and then a simple fully connected network with 128 hidden units for classification. We have used the SentenceTransformers Python library \cite{reimers2019sentencebert} and PyTorch to implement this loss function.

We have adapted BERT with Cross-Entropy Weighted Focal Loss (CEWF) \cite{wang2022bert} to our tasks by utilizing a pre-trained \texttt{bert-base-uncased} (110M parameters) model with mean pooling to produce sentence embeddings and then a simple fully connected neural network with 128 hidden units for classification. We have then used CEWF loss to train the classifier.

A second imbalanced-focused baseline is the Adaptable Focal Loss-based approach with BERT described in \cite{cao2022adaptable}. We have used the SentenceTransformers Python library \cite{reimers2019sentencebert} and PyTorch for this implementation. For this paper, we have benchmarked two implementations of the approach. In the first implementation, the fine-tuned network was completely disabled. In the second, the fine-tuning network was represented by a three-layer fully connected network, with the first and last layer consisting of 768 nodes (the embedding size of \texttt{bert-base-uncased}) and the middle layer consisting of 256 nodes. For the experimental evaluation, the encoder models were trained for 35 epochs on the training data.

As an example of a representation learning pipeline,
we provide a baseline building on the TSDAE architecture
\cite{wangTSDAEUsingTransformerbased2021}.

For the baseline in this paper, we have used the original version of TSDAE as implemented in the SentenceTransformers Python package by the authors. For the experimental evaluation, the encoder model was trained for $5$ epochs on the training data. Then, a 3-layer fully connected classifier was trained on the encoded embeddings of the training data using 5-fold cross-validation to determine the optimal size of the hidden layer. In our experiments, we start with a \texttt{bert-base-uncased} model pretrained using MLM and NSP objectives, and then further fine-tune it for sentence embeddings using the reconstruction loss.

Additionally, we also include pre-trained baselines. For this work, we have evaluated SimCSE \cite{gao2022simcse} (110M parameters), a contrastive approach to learning sentence embeddings. We also evaluate GPT-3.5 \cite{brown2020language} and GPT-4 \cite{gpt4} in a few-shot classification setting. For SimCSE, we have used the supervised pretrained uncased version of the model to encode both train and test sets. We have then followed the procedure indetical to the {\Modela} evaluation. For GPT-3.5 and GPT-4, we have sampled 5 samples of each class from the train set, and then queried the model for the class of samples from the test set. Due to cost conserns, we have only evalauted GPT-3.5 and GPT-4 on a stratified subset of the test set.

\subsection{TSDAE-{\Modela} Experimental Setup}

We initialize {\Modela} using the \texttt{bert-base-uncased} model weights. We then fine-tune the model on each dataset using the TSDAE procedure described above, using the {\Modela} loss instead of the reconstruction loss. Then, we compute embeddings for both the training and testing sets. Finally, a simple 3-layer perceptron is used to classify the embeddings. The amount of units in the hidden layer was optimized using grid-search cross-validation.

In our experimental implementation\footnote{Code and data used for the experiments will be made publicly available upon acceptance} of the {\Modela} loss function, both the batch size for the reconstruction loss and the $n$ (class sample size) parameter of {\Modela} were statically set to $3$. We did not perform any optimization of these parameters in our experiments. For the internal sampling of {\Modela}, we have used simple random sampling, separately taking samples from each class to ensure that the internal sample of the {\Modela} loss function remains balanced. In our experiments, we test the behavior of {\Modela} using three distance metrics, namely Euclidean, Chebyshev and cosine distances. 

For the experimental evaluation, the encoder
model was trained for five epochs on the training data. We have evaluated the model with different values of the $C$ parameter from 0 to 1. 

\subsection{Results and analysis}

\label{sec:exrs}

\begin{table*}
	[h]
	\centering
	\small
	
	\begin{tabular}{clrrr|clrrr}
		\toprule \textbf{DS}                                               & \textbf{Baseline}          & \textbf{F1}       & \textbf{PREC}     & \textbf{REC}      & \textbf{DS}                                              & \textbf{Baseline}          & \textbf{F1}       & \textbf{PREC}     & \textbf{REC}      \\
		\midrule \multirow{11}{*}{\rotatebox[origin=c]{90}{20 Newsgroups}} & TF-IDF+SVM                 & \underline{0.757} & \textbf{0.883}    & 0.662             & \multirow{11}{*}{\rotatebox[origin=c]{90}{India Police}} & TF-IDF+SVM                 & 0.764             & \underline{0.816} & 0.717             \\
		                                                                   & BoW+SVM                    & 0.707             & \underline{0.804} & 0.631             &                                                          & BoW+SVM                    & 0.775             & \textbf{0.822}    & 0.733             \\
		                                                                   & Doc2Vec+SVM                & 0.655             & 0.796             & 0.556             &                                                          & Doc2Vec+SVM                & 0.522             & 0.728             & 0.407             \\
		                                                                   & BERT + CE                  & 0.727             & 0.618             & 0.884             &                                                          & BERT + CE                  & 0.662             & 0.533             & 0.873             \\
		                                                                   & BERT + CEWF *              & 0.203             & 0.124             & 0.552             &                                                          & BERT + CEWF *              & 0.194             & 0.115             & 0.624             \\
		                                                                   & BERT Focal Loss w/ FT *    & 0.155             & 0.095             & 0.421             &                                                          & BERT Focal Loss w/ FT *    & 0.115             & 0.068             & 0.394             \\
		                                                                   & BERT Focal Loss w/o FT *   & 0.694             & 0.585             & 0.853             &                                                          & BERT Focal Loss w/o FT *   & 0.653             & 0.528             & 0.857             \\
		                                                                   & TSDAE                      & 0.704             & 0.579             & \textbf{0.900}    &                                                          & TSDAE                      & 0.658             & 0.517             & \underline{0.907} \\
		\cdashline{2-5} \cdashline{7-10}                                   & {SimCSE}                   & 0.745             & 0.654             & \underline{0.866} &                                                          & {SimCSE}                   & 0.693             & 0.582             & 0.857             \\
																		   & {GPT-3.5}                  & 0.429             & 0.300             & 0.750             &                                                          & {GPT-3.5}                  & 0.308             & 0.200             & 0.667             \\
		                                                                   & {GPT-4}                    & 0.571             & 0.667             & 0.500             &                                                          & {GPT-4}                    & \textbf{0.857}    & 0.750             & \textbf{1.000}    \\
		\cdashline{2-5} \cdashline{7-10}                                   & \textbf{{\Modela} $C=0.5$} & \textbf{0.790}    & 0.773             & 0.807             &                                                          & \textbf{{\Modela} $C=0.5$} & \underline{0.788} & 0.736             & 0.848             \\
		\midrule \multirow{11}{*}{\rotatebox[origin=c]{90}{BBC}}           & TF-IDF+SVM                 & 0.942             & \underline{0.961} & 0.925             & \multirow{11}{*}{\rotatebox[origin=c]{90}{InsightCrime}} & TF-IDF+SVM                 & 0.729             & \textbf{0.833}    & 0.648             \\
		                                                                   & BoW+SVM                    & 0.943             & 0.943             & 0.943             &                                                          & BoW+SVM                    & \underline{0.733} & 0.787             & 0.685             \\
		                                                                   & Doc2Vec+SVM                & 0.917             & 0.893             & 0.943             &                                                          & Doc2Vec+SVM                & 0.653             & 0.702             & 0.611             \\
		                                                                   & BERT + CE                  & \underline{0.955} & 0.914             & \textbf{1.000}    &                                                          & BERT + CE                  & 0.618             & 0.512             & \underline{0.778} \\
		                                                                   & BERT + CEWF *              & 0.347             & 0.238             & 0.642             &                                                          & BERT + CEWF *              & 0.253             & 0.168             & 0.519             \\
		                                                                   & BERT Focal Loss w/ FT *    & 0.201             & 0.124             & 0.528             &                                                          & BERT Focal Loss w/ FT *    & 0.332             & 0.214             & 0.741             \\
		                                                                   & BERT Focal Loss w/o FT *   & 0.929             & 0.881             & \underline{0.981} &                                                          & BERT Focal Loss w/o FT *   & 0.534             & 0.424             & 0.722             \\
		                                                                   & TSDAE                      & 0.912             & 0.852             & \underline{0.981} &                                                          & TSDAE                      & 0.571             & 0.452             & \underline{0.778} \\
		\cdashline{2-5} \cdashline{7-10}                                   & {SimCSE}                   & 0.895             & 0.836             & 0.962             &                                                          & {SimCSE}                   & 0.589             & 0.467             & 0.796             \\
		                                                                   & {GPT-3.5}                  & 0.588             & 0.417             & \textbf{1.000}    &                                                          & {GPT-3.5}                  & 0.303             & 0.179             & \textbf{1.000}    \\
		                                                                   & {GPT-4}                    & 0.600             & 0.600             & 0.600             &                                                          & {GPT-4}                    & 0.370             & 0.227             & \textbf{1.000}    \\
		\cdashline{2-5} \cdashline{7-10}                                   & \textbf{{\Modela} $C=0.5$} & \textbf{0.981}    & \textbf{0.964}    & \textbf{1.000}    &                                                          & \textbf{{\Modela} $C=0.5$} & \textbf{0.735}    & \underline{0.818} & 0.667             \\
		\midrule \multirow{11}{*}{\rotatebox[origin=c]{90}{CAMEO}}         & TF-IDF+SVM                 & 0.587             & 0.831             & 0.454             & \multirow{11}{*}{\rotatebox[origin=c]{90}{Spam}}         & TF-IDF+SVM                 & 0.940             & \underline{0.985} & 0.899             \\
		                                                                   & BoW+SVM                    & 0.564             & \underline{0.836} & 0.426             &                                                          & BoW+SVM                    & 0.915             & 0.963             & 0.872             \\
		                                                                   & Doc2Vec+SVM                & 0.155             & 0.476             & 0.093             &                                                          & Doc2Vec+SVM                & 0.880             & 0.926             & 0.839             \\
		                                                                   & BERT + CE                  & \underline{0.755} & 0.651             & \textbf{0.898}    &                                                          & BERT + CE                  & \underline{0.960} & 0.948             & \underline{0.973} \\
		                                                                   & BERT + CEWF *              & 0.374             & 0.261             & 0.657             &                                                          & BERT + CEWF *              & 0.211             & 0.125             & 0.664             \\
		                                                                   & BERT Focal Loss w/ FT *    & 0.392             & 0.269             & 0.722             &                                                          & BERT Focal Loss w/ FT *    & 0.125             & 0.076             & 0.342             \\
		                                                                   & BERT Focal Loss w/o FT *   & 0.625             & 0.518             & 0.787             &                                                          & BERT Focal Loss w/o FT *   & 0.856             & 0.773             & 0.960             \\
		                                                                   & TSDAE                      & 0.738             & 0.662             & 0.833             &                                                          & TSDAE                      & 0.956             & 0.959             & 0.953             \\
		\cdashline{2-5} \cdashline{7-10}                                   & {SimCSE}                   & 0.714             & 0.619             & 0.843             &                                                          & {SimCSE}                   & 0.919             & 0.888             & 0.953             \\
		                                                                   & {GPT-3.5}                  & 0.533             & 0.500             & 0.571             &                                                          & {GPT-3.5}                  & \textbf{1.000}    & \textbf{1.000}    & \textbf{1.000}    \\
		                                                                   & {GPT-4}                    & 0.706             & 0.600             & \underline{0.857} &                                                          & {GPT-4}                    & \textbf{1.000}    & \textbf{1.000}    & \textbf{1.000}    \\
		\cdashline{2-5} \cdashline{7-10}                                   & \textbf{{\Modela} $C=0.5$} & \textbf{0.860}    & \textbf{0.976}    & 0.769             &                                                          & \textbf{{\Modela} $C=0.5$} & \underline{0.966} & \underline{0.993} & 0.940             \\
		\bottomrule
	\end{tabular}
	\caption{Results of text classification experiments on all datasets. Methods marked
		with an asterisk represent imbalanced text classification approaches. \textbf{Bold}
		notations represent the best dataset-level results per metric. The second-best
		results are \underline{underlined}. For {\Modela}, the $C$=0.5 value was chosen
	as a balanced default value.}
	\label{ExpResultsTable}
\end{table*}

\begin{figure*}[!h]
    \centering
    \includegraphics[width=\textwidth]{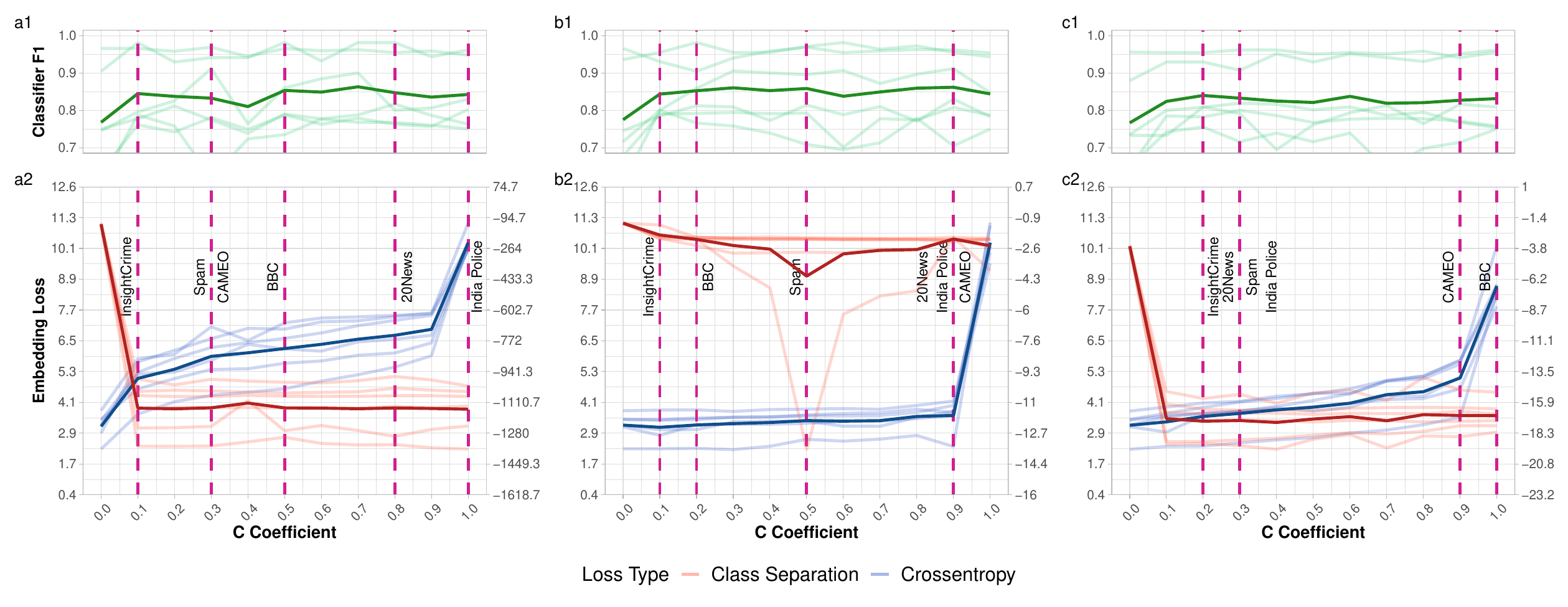}
    \caption{Relationship between classifier F1 (1), different loss components (class separation and reconstruction) (2), and the value of $C$, averaged across datasets, for three different distance measures: Euclidean (a), Cosine (b), Chebyshev (c).}
	\label{CValueStudy}
\end{figure*}

The results of the experiments are presented in Table \ref{ExpResultsTable}. For consistency in comparison to baselines and across the datasets, results reported in Table \ref{ExpResultsTable} use $C=0.5$, balancing the influence of the reconstruction and class separation losses. Overall, they show that \Modela-based optimization leads to a marked increase in
classification performance in under-represented classes across most of the tested datasets. The only datasets where the proposed pipeline had performance close to that of the baseline methods were the India Police and InsightCrime datasets. On average, {\Modela} shows a $4.1$\

Finally, we can point out that our model performs at least as well in terms of precision as in terms of recall (precision is larger than recall on three datasets). This indicates that the proposed model tends to correctly identify the instances of the minority class, even in the presence of adverse conditions.

\subsection{Optimal $C$ value experiments}

\begin{table*}[!h]\centering
\small

\begin{tabular}{lccc|ccc|ccc|ccc}\toprule
\multirow{2}{*}{\textbf{Dataset}} &\multicolumn{3}{c}{\textbf{C=0}} &\multicolumn{3}{c}{\textbf{C=0.5}} &\multicolumn{3}{c}{\textbf{C=1}} &\multicolumn{3}{c}{\textbf{Optimized C}} \\\cmidrule{2-13}
&\textbf{F1} &\textbf{SI} &\textbf{kDN} &\textbf{F1} &\textbf{SI} &\textbf{kDN} &\textbf{F1} &\textbf{SI} &\textbf{kDN} &\textbf{F1} &\textbf{SI} &\textbf{kDN} \\\midrule
20News &0.749 &\underline{0.935} &0.082 &\underline{0.790} &\underline{0.935} &\underline{0.073} &0.763 &0.916 &0.089 &\textbf{0.817 (at C=0.8)} &\textbf{0.936} &\textbf{0.067} \\
BBCNews &0.904 &0.975 &0.035 &\textbf{0.981} &\underline{0.981} &\textbf{0.009} &\underline{0.962} &\textbf{0.988} &\underline{0.015} &\textbf{0.981 (at C=0.5)} &\underline{0.981} &\textbf{0.009} \\
CAMEO &0.745 &0.820 &0.204 &\underline{0.860} &0.931 &0.067 &0.830 &\textbf{0.942} &\underline{0.059} &\textbf{0.912 (at C=0.3)} &\underline{0.938} &\textbf{0.058} \\
IndiaPoliceSent &0.635 &0.892 &0.120 &\underline{0.788} &\underline{0.937} &\underline{0.061} &\textbf{0.802} &\textbf{0.949} &\textbf{0.054} &\textbf{0.802 (at C=1.)} &\textbf{0.949} &\textbf{0.054} \\
InsightCrime &0.607 &0.862 &0.173 &0.735 &\underline{0.912} &\underline{0.100} &\underline{0.750} &0.909 &0.102 &\textbf{0.787 (at C=0.1)} &\textbf{0.928} &\textbf{0.097} \\
Spam &\underline{0.966} &0.926 &0.097 &\underline{0.966} &\underline{0.987} &\underline{0.014} &0.948 &0.986 &0.015 &\textbf{0.969 (at C=0.3)} &\textbf{0.988} &\textbf{0.013} \\
\bottomrule
\end{tabular}
\caption{F1 and class overlap measures for testing datasets and corresponding values of the $C$ coefficient per dataset (line). \textbf{Bold} notations represent best dataset-level results per metric. The second-best results are \underline{underlined}.}
\label{tab:ablst}
\end{table*}

To test the effect of the $C$ regularization value on the model's behavior, we have run the pipeline on the
datasets used for testing with different values of $C$, then compared the final classifiers' performance. We can see the relationship between the final classifier F1, as well as different loss values and the
value of $C$ in Figure \ref{CValueStudy}. Once the class separation loss starts being used for the optimization ($C>0$), it begins to decrease, while the reconstruction loss begins to increase. The reconstruction loss peaks when $C=1$, as when $C=1$, it is not used for optimization.

Based on Figure \ref{CValueStudy}, we can see that $C$ is domain-dependent, with optimal values of $C$ (marked in purple) being different for different datasets. We can also see the relationship between the different loss components and $C$. From this, we can conclude that the optimal value of $C$ is domain-dependent and connected with certain dataset features used for model training and testing.

\begin{figure*}[!ht]
     \centering
    \includegraphics[width=\textwidth]{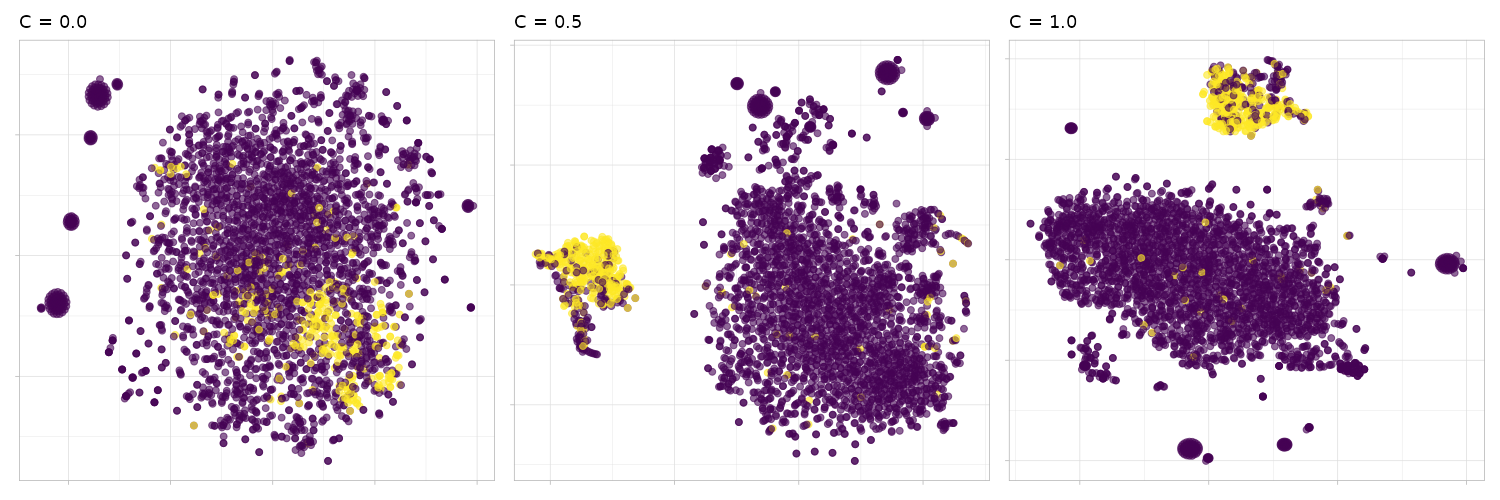}
    \caption{Comparison of TSDAE-{\Modela} embeddings with $C=0$, $C=0.5$  and $C=1.0$ (India Police dataset)}
    \label{VanillaCMPIP}
\end{figure*}

We can also see that the embeddings computed by the model trained using the TSDAE-{\Modela} loss (Figure \ref{VanillaCMPIP}) show a clearer decision boundary than the embeddings computed using the original TSDAE pipeline trained on the same dataset ($C=0$). 

\subsection{Ablation study}

In this section, we compare different extreme settings for the regularization parameter $C$ to better understand our model's behavior under different conditions and the effect that the $C$ parameter has on the model's behavior. 
Additionally, and as a demonstration, we have computed visualizations of the embeddings of the test set of the India Police dataset, as this was one of the datasets where the proposed model did not outperform the baseline
methods. To visualize the embeddings, we have used t-SNE
to create a 2-dimensional representation of the embeddings.

We computed class overlap measures of the test set embeddings of all the datasets used in our experiments. We use the Separability Index (SI) measure, the proportion of the points with the closest neighbor of the same class. We also use the K-Disagreeing neighbors measure (kDN), which is the average proportion of neighbors of a different class for each of the points (for our computation, we have set $k=5$) \cite{cOvelapMeasures}.

The extreme values of the $C$ parameter were chosen because they allow us to compare the performance of $\Ls_{\Modela}$ to its components operating independently. First, we can look at the model when $C=0$, which is analogous to the baseline TSDAE pipeline that does not use class separation.
As we can see in Figure \ref{VanillaCMPIP}, this can result in a large class overlap in the embedding space, decreasing classification performance. 

Next, with $C=1$, the optimizer only focuses on class separation. With this, the optimization process does not pay attention to reconstruction, leading to decreases in classification performance (Figure~\ref{CValueStudy} and Table~\ref{tab:ablst}), as the embeddings start to lose meaning with the reconstruction loss not included in the optimization.

With $C=0.5$, the optimization process gives equal attention to the class separation and reconstruction
losses. This allows the optimization process to maintain reconstruction ability while separating classes in the embedding space. However, this does not often maximize classification performance, as evidenced by Figure \ref{CValueStudy} and Table \ref{tab:ablst}. We believe
this is because, for some datasets (e.g., InsightCrime), classes are more separated by the reconstruction loss function in the embedding space. Thus, focusing more on reconstruction can lead to optimal performance. The converse can be true for other datasets, such as IndiaPolice, where classes are very mixed in the embedding space, and thus, focusing more on class separation
can further improve performance.

This leads us to conclude that, considering the domain dependence of the optimal $C$ values, care should be taken when optimizing the value of $C$ for a particular domain-specific dataset to achieve maximum performance.

\section{Conclusion}
\label{sec:co}

In this paper, we address the problem of imbalanced text classification by introducing a novel loss function, {\Modela}. {\Modela} is designed to produce embeddings where classes are separated while maintaining the
semantic value of the embeddings produced by models trained with this loss function. The proposed loss function allows the researchers to determine the trade-off level between the embeddings' faithfulness and class separation. We also show that the level of trade-off between reconstruction and class separation is domain-dependent and that the optimal value of $C$ depends on the characteristics of a particular dataset. We also show that, as with all other regularization techniques, careful consideration of the $C$ value is needed, as an incorrect choice can decrease model performance. This allows {\Modela} to create high-quality sentence embeddings in highly imbalanced scenarios, thus improving the performance of downstream tasks. For example, {\Modela} embeddings show high performance for text classification tasks on various datasets, faring better than other models aimed at imbalanced text classification. Moreover, we show that {\Modela} can produce embeddings
with a clear separation between classes, enabling the use of this model for downstream tasks other than text classification.

\section{Limitations}
The first limitation of this work is that the performance of the model was only evaluated on English-language datasets. While we believe that the {\Modela} will transfer well to other languages, this claim needs to be evaluated separately. The use of the loss function to fine-tune on English-language datasets does not necessarily facilitate an increase of performance in other languages. We have also shown {\Modela} to be domain sensitive, which means that models trained on one domain should not be applied to other domains without careful evaluation.

We can also see that even in the embeddings generated with the {\Modela} loss, some of the data points were still closer to the incorrect class in the embedding space (e.g. \ref{VanillaCMPIP}). This requires further investigation, as these mis-embeddings can point out the potential weaknesses of the proposed method. In practice, this means that the model should be carefully evaluated for it's applicability for a particular domain before being deployed to production, especially in domains where misclassifications pose a significant risk. Additionally, {\Modela} still requires a large amount of human-annotated data to train. The method is also sensitive to the $C$ hyperparameter, which means that cross-validation or other extensive testing will be needed before using models trained with {\Modela} in production. 

Finally, an important limitation of this work is that we have only evaluated the performance of our method on a binary classification task. Our method can be extended to more classes, but this will lead to increases of the computational complexity, which is why we decided to relegate this possible extension of the method to future work.

\bibliographystyle{ACM-Reference-Format}
\bibliography{ML-Bibliography}

\end{document}